\documentclass[runningheads]{llncs}
\usepackage[T1]{fontenc}
\usepackage{graphicx}
\usepackage{booktabs}
\usepackage[misc]{ifsym}

\usepackage{algorithm}
\usepackage{algorithmic}
\usepackage{amsfonts}
\usepackage{amsmath}
\usepackage{color}
\usepackage{comment}

\newcommand{\Rb}{\mathbb{R}}

\newcommand{\Ic}{\mathcal{I}}

\DeclareMathOperator*{\argmax}{arg\,max}

\begin{document}

\title{Select-to-Act: Hierarchical Reinforcement Learning via Adaptive Language Guidance}

\titlerunning{Select-to-Act: Hierarchical RL via Adaptive Language Guidance}

\author{Hanping Zhang\inst{1} \and Adam Koziak\inst{1} \and Yuhong Guo\inst{1,2}}

%
\authorrunning{H. Zhang et al.}


\institute{School of Computer Science, Carleton University, Ottawa, Canada 
\and Canada CIFAR AI Chair, Amii, Canada\\
\email{\{jagzhang@cmail., adamkoziak@cmail., yuhong.guo@\}carleton.ca} 
}

\toctitle{Select-to-Act: Hierarchical Reinforcement Learning via Adaptive Language Guidance}

\tocauthor{Hanping Zhang, Adam Koziak, Yuhong Guo}

\maketitle

\begin{abstract}
Reinforcement Learning (RL) has been widely applied to sequential decision-making, 
yet it often suffers from poor sample efficiency due to costly interactions with the environment. 
A limited line of recent work has started exploring improving RL efficiency 
by leveraging external knowledge expressed in natural-language instructions. 
However, the few existing
approaches typically treat the entire instruction as a single conditioning input, 
failing to account for the stage-dependent nature of language guidance,
especially in complex environments.
In this paper, we propose \emph{Hierarchical Reinforcement Learning with Language Instructions (HRLLI)}, 
a hierarchical RL framework that explicitly models natural-language instructions 
as dynamically selectable semantic guidance during decision-making. 
HRLLI decomposes instructions into a set of piecewise guidance elements, 
where each instruction piece may become relevant at different stages of interaction with the environment. 
A novel hierarchical RL policy structure is then formulated in a \emph{Select-to-Act} paradigm: 
a high-level semantic policy acts as a guidance selector 
that selects the most relevant instruction piece to the current state to guide the low-level agent's decision,
while a low-level policy executes environment actions conditioned on the selected guidance. 
The two-level policies are learned simultaneously to maximize augmented expected returns 
from interactions with the environment.
This design enables the agent to adaptively ground language instructions into stage-specific 
decisions during interaction.
Experiments on the instruction-intensive RTFM benchmark show that HRLLI consistently outperforms strong instruction-conditioned RL baselines, demonstrating that explicitly 
modeling adaptive instruction selection significantly improves the effectiveness of RL. 
\end{abstract}
%
\section{Introduction}
Reinforcement learning (RL), as a powerful framework for sequential decision making,
has been successfully applied in many domains, including robotics~\cite{tang2025deep}, video games~\cite{berner2019dota}, autonomous driving~\cite{kiran2021deep}, and even large language model (LLM) training~\cite{ouyang2022training}. Despite these successes, RL often suffers from poor sample efficiency due to the large number of interactions required with the environment, particularly in long-horizon or sparse-reward tasks~\cite{sutton1998reinforcement}.
Prior works have considered improving the sample efficiency of RL by incorporating additional sources of knowledge beyond reward signals, such as offline demonstration data~\cite{pertsch2021accelerating,zhangECAI25} 
and human feedback~\cite{hu2023language}. Compared with carefully designed reward functions, such auxiliary data are often easier to obtain in many real-world scenarios.

In many realistic settings, such auxiliary knowledge is often available 
in natural language~\cite{luketina2019survey}.
Prior studies have explored various ways of integrating language into RL, 
including using human language feedback~\cite{hu2023language}, 
full textual descriptions of the environment~\cite{shridhar2020alfworld}, or a single sentence of language descriptions of tasks to facilitate policy learning for complex robotic manipulation tasks~\cite{mees2022calvin} and improve generalization across multiple tasks~\cite{yu2020meta}. 
While language has been widely explored in RL, \emph{language-instructed RL}—which leverages textual instructions that provide high-level semantic guidance, or piecewise descriptions of how a task can be completed—remains largely underexplored, with only a very limited number of prior works~\cite{zhong2019rtfm,dainese2023reader}. 

This setting is particularly interesting and practical in real-world scenarios, 
where humans can often easily provide a set of piecewise tips for completing a task 
rather than a single long sentence describing the task or a fully specified step-by-step instruction sequence. 
In such cases, each instruction tip in the set may only be relevant at a particular stage of exploration rather than throughout the entire interaction. 
However, prior work has not yet explicitly formalized such an instruction set, 
nor studied how to effectively utilize the knowledge contained in it. 
Existing language-instructed RL approaches typically encode all instructions together and condition the policy on an aggregated instruction representation~\cite{dainese2023reader,zhong2019rtfm}, 
implicitly assuming that all instructions are equally relevant at every stage of decision making. 
This assumption overlooks the stage-dependent nature of instruction usefulness, where different instruction pieces may become relevant at different phases of interaction with the environment. 
As a result, encoding the entire instruction set without distinguishing their relevance can lead to ineffective language grounding and limit the potential benefits of language guidance for improving RL training efficiency.

In this paper, we formally formulate the setting of language-instructed RL, where, in addition to the standard environment dynamics and reward signals, a set of natural language instructions is provided to guide the agent. Each instruction can offer high-level semantic guidance that may become relevant at different stages of task completion. To address this problem, we propose \textbf{H}ierarchical \textbf{R}einforcement \textbf{L}earning with \textbf{L}anguage \textbf{I}nstructions (HRLLI), a hierarchical selector–executor framework that learns separate policies for instruction selection and action execution. At the high level, an instruction policy acts as a 
\emph{selector} that chooses the most relevant instruction based on the current state observation. 
At the low level, an action policy acts as an \emph{executor} that follows the selected instruction to produce actions for several environment steps. This hierarchical design allows the agent to effectively utilize guidance from the instruction set while improving interpretability compared to traditional instruction-conditioned policies. We evaluate HRLLI on the instruction-heavy RTFM benchmark~\cite{zhong2019rtfm}. 
Experimental results show that HRLLI consistently outperforms strong instruction-conditioned baselines, 
$\text{txt}2\pi$~\cite{zhong2019rtfm} and Reader~\cite{dainese2023reader}, across multiple environment settings.
Our main contributions are summarized as follows:
\begin{itemize}
\item We study an underexplored setting of language-instructed RL in which the agent receives an instruction set consisting of multiple pieces of guidance whose relevance depends on the given state.
\item We propose HRLLI, a novel hierarchical select-to-act framework for language-instructed RL 
	that decomposes instruction usage into a high-level instruction selection policy and 
	a low-level instruction-guided action execution policy.
\item We conduct experiments on the instruction-heavy RTFM benchmark across multiple settings,
	demonstrating that HRLLI consistently outperforms multiple state-of-the-art language-instructed RL baselines.
\end{itemize}
%
\section{Related Work}
\subsection{Reinforcement Learning with Language as Auxiliary Information}
To improve the training efficiency of RL, many prior works have explored incorporating natural language as auxiliary information during training. 
Luketina et al.~\cite{luketina2019survey} provide a comprehensive survey of RL methods that integrate natural language, categorizing them into two main types: language-conditional RL, where natural language is essential to policy learning and forms part of the problem formulation, and language-assisted RL, where language is used to facilitate RL training but is not strictly required for solving the task.
Following this line of work, a variety of benchmarks and approaches have been developed to incorporate language as auxiliary guidance in RL. 
Mees et al.~\cite{mees2022calvin} introduced the CALVIN challenge, which provides a natural-language description of each robotic manipulation task alongside complex environment interactions, enabling agents to learn manipulation skills more efficiently.
Similarly, Meta-World provides language descriptions of tasks as auxiliary information within the environment~\cite{yu2020meta}. Unlike CALVIN, Meta-World consists of a diverse set of tasks, each associated with a short language description, allowing agents to learn meta-RL policies that can generalize or rapidly adapt across tasks.
ALFWorld~\cite{shridhar2020alfworld} and ALFRED~\cite{shridhar2020alfred} are two closely related benchmarks that provide textual task descriptions and align state observations and actions with natural language, facilitating the integration of language into RL learning.
With the rapid development of LLMs, language information can be incorporated into RL in more sophisticated ways. 
For example, Wang et al.~\cite{wang2024llm} proposed the LLM-Empowered State Representation (LESR) method, which leverages LLMs to generate code for constructing task-relevant state representations that improve RL training efficiency. 
Yan et al.~\cite{yan2024efficient} further proposed incorporating language inputs into an LLM and using the resulting LLM policy as a prior to guide RL policy learning.

\subsection{Language-instructed Reinforcement Learning}
Although incorporating language as auxiliary information for RL has been widely studied, using natural language as explicit instructions to guide RL training remains relatively underexplored. 
Zhong et al.~\cite{zhong2019rtfm} initiated the study of language-instructed RL and proposed the Read to Fight Monsters (RTFM) environment, which provides textual instructions describing how to complete tasks alongside standard environment dynamics, serving as a benchmark for studying language-instructed RL.
They also proposed the $\text{txt}2\pi$ method to address the RTFM challenge.
Subsequent work by Dainese et al.~\cite{dainese2023reader} proposed REinforcement learning Agent for Discrete Environments with wRitten instructions (Reader), which builds a world model in the RTFM environment to learn an instruction-conditioned reward model and then applies Monte Carlo Tree Search (MCTS)~\cite{kocsis2006bandit} for planning.
More recent studies have explored related directions of language-instructed RL in different settings. 
Hu and Sadigh~\cite{hu2023language} introduced InstructRL, which aligns RL policies with human preferences using natural language instructions, although their approach relies heavily on real human interactions. 
Baek et al.~\cite{baek2025ipcgrl} proposed Instruction-based Procedural Content Generation via RL (IPCGRL), which leverages language instructions to guide procedural level generation for RL training. 
However, their focus is on procedural environment generation rather than instruction-guided decision making, making it fundamentally different from our setting and thus not directly comparable.
%
%
\section{Problem Setting}
\label{sec:problem}
We consider an RL setting where the agent is provided with 
additional knowledge in the form of language instructions. 
Formally, we model the problem as an instruction-augmented version of the standard Markov Decision Process (MDP)~\cite{sutton1998reinforcement}, defined as $\mathcal{M}=(\mathcal{S},\mathcal{A},\mathcal{T},\mathcal{R},\gamma,\mathcal{I})$, where $\mathcal{S}$ and $\mathcal{A}$ denote the state and action spaces, $\mathcal{T}:\mathcal{S}\times\mathcal{A}\rightarrow\mathcal{P}(\mathcal{S})$ represents the transition dynamics, $\mathcal{R}:\mathcal{S}\times\mathcal{A}\rightarrow\mathbb{R}$ is the reward function, 
$\gamma\in(0,1)$ is the discount factor, 
and $\mathcal{I}$ denotes the additional natural-language instruction set. 

\begin{figure}[t]
\centering
\includegraphics[width=0.9\linewidth]{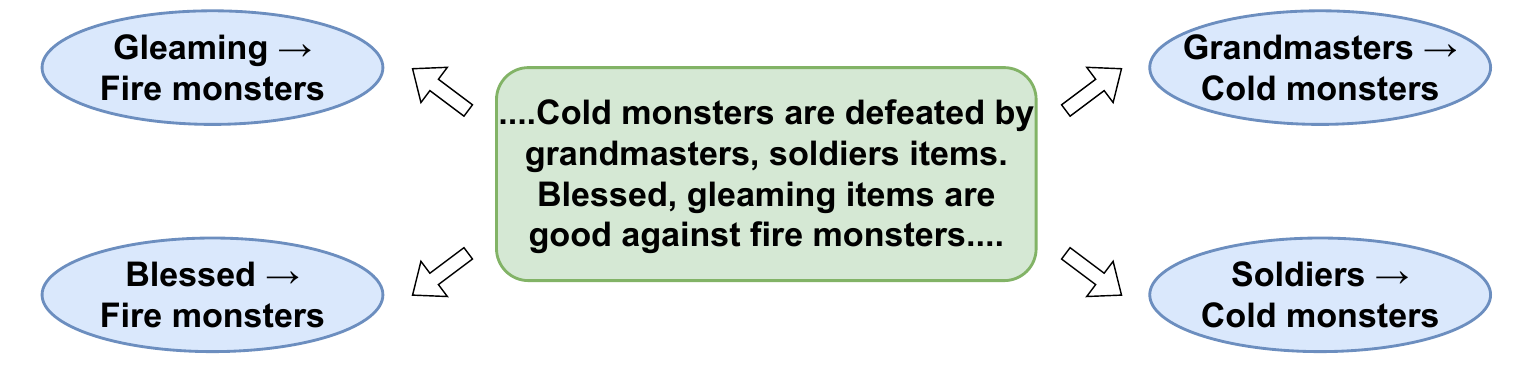}
\caption{Example of an instruction set in RTFM environment.}
\label{fig:instruction}
\end{figure}

The textual instructions in $\mathcal{I}$ 
provide guidance information on how to complete the target task of RL. 
We assume that the instruction can be decomposed into a set of semantic tips, 
such as $\mathcal{I}=\{I_1,\dots,I_{|\mathcal{I}|}\}$, where the instruction set has variable length $|\mathcal{I}|$. Each element $I_i$ is a sentence-level textual description that provides useful,
often stage-dependent guidance for accomplishing the task. 
Importantly, these descriptions typically do not form a complete step-by-step procedure for solving the task. Instead, each tip provides short-term guidance for achieving a subgoal of the task (e.g., Figure~\ref{fig:instruction}). Such a formulation is common in real-world scenarios, where tasks are rarely specified as fully detailed procedures but are instead described through a collection of partial hints or useful tips. 
The instruction set is accessible throughout the RL 
as auxiliary information for decision making. 
The objective is to learn an optimal policy $\pi^\star$ that maximizes the expected long-term discounted return 
by exploiting the instruction set.
%
%
\section{Method}
In this section, we present our proposed Hierarchical Reinforcement Learning with Language Instructions (HRLLI) framework 
to tackle language-instructed RL. 
HRLLI exploits the high-level semantic knowledge in the instruction set to improve RL
through a novel hierarchical policy learning paradigm. 
At the high level, both instructions and current state observations are encoded into a shared embedding space, where each instruction is scored with respect to the given state. 
The high-level policy then selects the most relevant instruction, which serves as 
guidance for the next few timesteps of low-level action execution. 
At the low level, the action policy takes the selected instruction as additional contextual information and outputs actions to interact with the environment. 
An overview of the overall framework is illustrated in Figure~\ref{fig:overview},
while the design philosophy and methodological details are presented below. 
%
%
\subsection{Design of the Hierarchical Select-to-Act Paradigm}
\begin{figure}[t]
\centering
\includegraphics[width=0.98\linewidth]{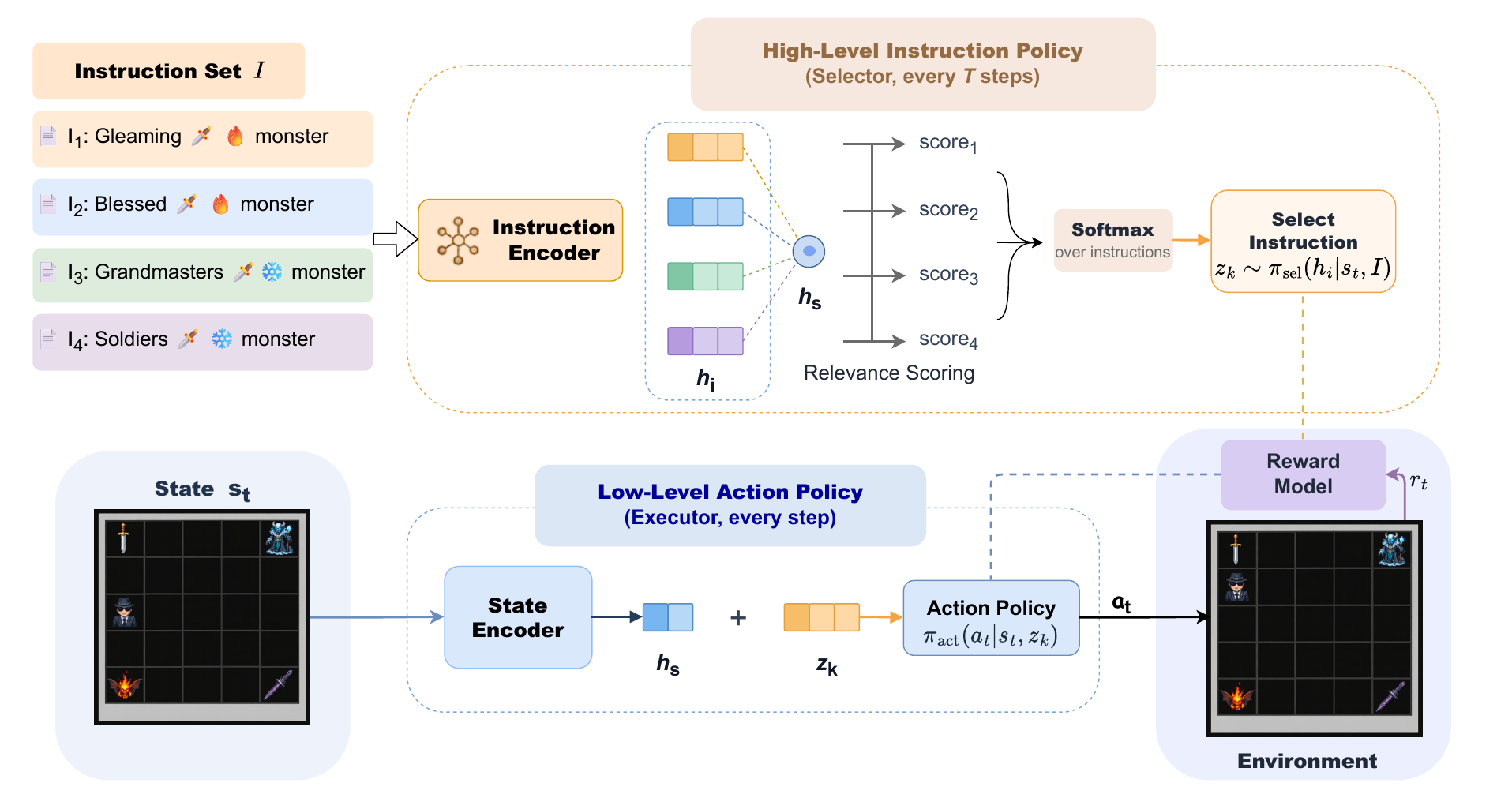}
\caption{
Overview of our hierarchical instruction-based RL framework. At every high-level decision point, the selector encodes the instruction set and the current state into a shared latent representation space, computes relevance scores for all instructions, and selects an 
instruction from a softmax distribution, which is used to produce
the guidance vector $z_k$. For the subsequent $T$ environment steps, the low-level executor conditions on both the current state embedding and $z_k$ to take actions. A reward model is learned to provide dense shaping signals to facilitate policy training under sparse environment rewards.
}
\label{fig:overview}
\vskip -0.1in
\end{figure}
Under the language-instructed RL setting formalized in Section~\ref{sec:problem}, 
the instructions in $\mathcal{I}$ serve as stage-dependent semantic guidance.
Consequently, under a given state, only a subset of instructions may be relevant,
while the remaining ones may be irrelevant or even distracting.
In such cases, consuming the entire instruction set and conditioning the policy
on a single aggregated representation (e.g., via pooling or attention), as in prior works,
is undesirable. Such aggregation can blur the semantics of individual instruction tips
and make it more difficult for the policy to infer useful guidance at the current
stage of interaction.

Motivated by this observation, we view the language-instructed RL problem
as a dual decision-making problem: selecting the relevant instruction from a
variable-length instruction set and selecting the corresponding action to execute
in the environment.
However, these two decisions operate at different temporal scales and require
different learning signals.
Instruction selection is a discrete decision whose effect should be evaluated
at the segment level (i.e., over an action sequence), whereas action control
requires fine-grained step-level feedback.
Optimizing both decisions under a single objective with sparse rewards is therefore difficult,
as the policy must simultaneously infer the relevant instruction and execute actions,
leading to unstable credit assignment.

To address this challenge, we propose a bi-level hierarchical policy learning
paradigm, HRLLI, that learns two interdependent policies with distinct objectives.
A high-level instruction policy, acting as a guidance selector, operates at a
slower timescale and only reselects the instruction 
most relevant to the current state for every $T$ environment timesteps,
a low-level action policy, acting as an executor, performs action selection
conditioned on the selected instruction for the next $T$ timesteps.
Correspondingly, the selector is trained using segment-level outcomes from $T$ timesteps,
whereas the executor is optimized using step-level reward feedback.
This design aligns the optimization objectives with the roles of the two policies,
resulting in more stable learning and credit assignment.

%
\subsection{High-Level Instruction Policy: Selector}
For the high-level policy, we aim to learn a selector that can choose the most appropriate instruction for the current state $s_t$. 
To enable state-dependent selection, we embed
instructions and states in a shared latent representation space, so the selector can measure their compatibility directly, 
allowing the high-level policy to better align the two modalities and 
select instructions that are informative for the current state.
%
\paragraph{\bf Instruction and State Encoders}
Given the instruction set $\Ic=\{I_1,\dots,I_{|\Ic|}\}$, each instruction $I_i$ is first processed by a pretrained language model (LM). We adopt MiniLM~\cite{wang2020minilm} as our base model due to its efficiency and strong semantic representations. Directly fine-tuning a large LM within RL is computationally expensive and can destabilize training because RL gradients are noisy and non-stationary. Therefore, we freeze the base LM and only learn a lightweight MLP projection head as an adaptor, parameterized by $\phi_I$. This adaptor serves two purposes: (i) it maps LM features into the task-specific latent space used by the RL agent; and (ii) it provides sufficient capacity to align language embeddings
with state representations without overfitting the LM.
Specifically, 
the embedding vector $h_i^{I}$ for instruction $I_i$ is computed as:
\begin{equation}
  h_i^{I} = f_{\text{inst}}^{\phi_I}(I_i)\in\Rb^d,
	\qquad\mbox{where}\quad
	f_{\text{inst}}^{\phi_I}(\cdot) = \mathrm{MLP}_{\phi_I}(\mathrm{MiniLM}(\cdot)).
\end{equation}
Similarly, the state is encoded through a state encoder $f_\text{state}^{\phi_S}$
parameterized by $\phi_S$. 
Following the language-conditioned residual CNN architecture in $\text{txt2}\pi$~\cite{zhong2019rtfm}, 
the state encoder maps a state $s_t$ to a $d$-dimensional state embedding $h_t^S$:
\begin{equation}
  h_{t}^{S} = f_{\text{state}}^{\phi_S}(s_{t})\in\Rb^d .
\end{equation}
Using a shared dimension $d$ 
allows the selector to compare state and instruction embeddings with a simple compatibility function. 
Moreover, the same state encoder is used in the low-level executor policy learning as well. 
The state encoder is updated using gradients from both the selector and executor objectives. 
This shared training facilitates effective
state embeddings to preserve information useful for 
both instruction grounding (high-level) and action execution (low-level). 

\paragraph{\bf Instruction Selection Policy}
As completing a subtask in an RL environment typically requires a short sequence of actions, 
a high-level semantic instruction tip can often provide informative guidance 
for executing such an action sequence. 
For simplicity, we introduce a horizon $T$ to denote the fixed influential range 
of a selected instruction over sequential action executions. 
That is, the high-level instruction is reselected every $T$ environment timesteps. 
When $T=1$, the problem degenerates into a simple setting where the instruction is reselected at every 
environment timestep, which may lead to lower efficiency as well as potential confusion and inconsistency 
in selecting and executing instruction guidance.
Therefore, the $k$-th high-level instruction selection step
corresponds to the $kT$-th low-level environment timestep.
At this point, an instruction is selected based on the current state embedding $h_{kT}^S$.
Specifically, at each selection step $k$, we first compute the compatibility scores between
the state embedding and each instruction embedding as follows:
\begin{equation}
  \ell_{k,i} = \mathrm{score}_{\theta_H}\left(h_{kT}^{S}, h_i^{I}\right), 
	\; \forall i\in\{1,2,\cdots, |\mathcal{I}|\},
\end{equation}
where $\mathrm{score}_{\theta_H}(\cdot,\cdot)$ is a learnable scoring function
parameterized by $\theta_H$. 
Intuitively, $\ell_{k,i}$ captures the compatibility between the current 
state $s_{kT}$ and the semantic content of instruction $I_i$.
Based on the compatibility scores, the instruction selection policy 
is computed using the following softmax function:
\begin{equation}
  \pi_{\text{sel}}(i|s_{kT},\Ic)
  =
  \frac{\exp(\ell_{k,i})}{\sum_{j=1}^{|\Ic|}\exp(\ell_{k,j})}.
\end{equation}
The softmax converts scores into a normalized categorical distribution over the instructions in set $\mathcal{I}$. 
Therefore, during training we sample an instruction from the resulting categorical distribution, i.e.,
\begin{equation}
  i_k^* \sim \pi_{\text{sel}}(\cdot|s_{kT},\Ic),
\end{equation}
which enables stochastic exploration while stabilizing learning. 
During evaluation, we adopt a hard selection rule by choosing the most relevant instruction 
with the highest score,
\begin{equation}
  i_k^*=\argmax_{i\in[1,2,...,|\mathcal{I}|]}\ell_{k,i},
\end{equation}
ensuring deterministic and consistent instruction selection in evaluation.

To facilitate low-level policy learning, we map the selected instruction embedding to a guidance vector.
Since the instruction embeddings are already represented in the same embedding space 
as the state representations,
we empirically find that directly using the selected instruction embedding 
is sufficient and does not require additional transformations:
\begin{equation}
  z_k = g(h_{i_k^*}^{I}) = h_{i_k^*}^{I}.
\end{equation}
This choice avoids introducing an extra transformation that may distort the semantic geometry of pretrained language embeddings.
%
\paragraph{\bf Training Objective of High-Level Policy}
As each selected instruction guides the action selection 
of the low-level executor for the next $T$ environment timesteps,  
the selector should prefer instructions that lead to better reward outcomes 
from its guided episode segment. 
Accordingly, we define the following segment return 
to align credit assignment with instruction selection: 
\begin{equation}
  R_k = \sum_{t=0}^{T-1} r_{kT+t},
\end{equation}
which aggregates rewards over the next $T$-step episode segment after committing to the selected instruction
at selection step $k$.

However, environment rewards in language-instructed benchmarks are often sparse and delayed. If the selector is trained only from $R_k$, learning can be very slow because many segments may receive zero reward even if the selector makes correct decisions. 
To provide denser learning signals, 
we deploy a learnable reward model $\hat{r}_\psi$ 
to compute a predicted segment return $\hat{R}_k = \sum_{t=0}^{T-1}\hat{r}_{kT+t}$. 
This is expected to improve training efficiency by providing informative gradients even when the environment reward is sparse.
Even so, since the reward model is learned from environment feedback, 
its quality depends heavily on the performance of the low-level executor. 
Early in training, when the executor is still immature, segment reward outcomes are often noisy, 
which can lead the reward model to provide inaccurate estimates and consequently misguide the selector.
To address this issue, we further introduce an auxiliary reward that directly measures the alignment between the selected instruction and the trajectory segment it induces. 
Specifically, the auxiliary reward is defined as the cosine similarity between the embedding of the selected instruction and the average state embedding over the corresponding segment. 
This design is motivated by the shared latent representation space: when an instruction is relevant to the agent's behavior, the states visited while following that instruction should produce embeddings that are well aligned with the instruction embedding.
Following this definition, 
we compute the average state embedding over the segment as
\begin{align}
  u_k = \frac{1}{T}\sum_{t=kT}^{(k+1)T-1} h_t^{S},
\end{align}
and compute the auxiliary reward as: 
\begin{align}
  R_k^{\text{aux}} = \cos(u_k, z_k).
\label{eqa:auxiliary}
\end{align}
By providing a dense and low-variance training signal, this auxiliary reward encourages 
consistency between the selected instruction and the induced trajectory, 
thereby stabilizing selector learning in environments with sparse feedback.

By integrating both the predicted segment-reward and the auxiliary reward, 
we define the selector training objective as 
\begin{equation}
  J_{\pi_{\text{sel}}}(\theta_H,\phi_S,\phi_I) = \mathbb{E}_{\tau\sim \pi_{\text{sel}}}
  \left[
    \sum_{k=0}^{K-1}\gamma_H^k \left(\hat{R}_k + \lambda_\text{sel} R_k^{\text{aux}}\right)
  \right],
  \label{eqa:high-level}
\end{equation}
where $\gamma_H$ is the discount factor at the instruction level. 
We optimize the high-level policy by maximizing this objective using the REINFORCE algorithm~\cite{sutton1999policy}, which provides a simple and effective optimization strategy for this setting.
%
%
\subsection{Low-Level Action Policy: Executor}
The selected instruction represents a subtask-specific guidance signal. 
We condition the low-level action policy $\pi_{\text{act}}$ 
on the selected instruction embedding $z_k$, 
guiding the low-level decision making for the next $T$ timesteps with the selected instruction. 
At each timestep $t\in[kT,(k+1)T-1]$, 
the low-level executor samples an action as
\begin{equation}
a_t \sim \pi_{\text{act}}(\cdot|h_t^{S}, z_k;\theta_L),
\end{equation}
where $\theta_L$ denotes the parameters of the executor network. 
Conditioning the executor on $z_k$ provides additional semantic context beyond the raw state, helping the policy disambiguate which behaviors are currently relevant. 
Without this conditioning, the executor would need to infer the intended subtask solely from environment observations,
significantly increasing the difficulty of policy learning.
Moreover, conditioning on a single selected instruction instead of an aggregated instruction embedding simplifies policy learning: the executor learns a consistent mapping from $(s_t,z_k)$ to actions, 
where $z_k$ remains fixed for $T$ timesteps. This stabilizes exploration 
and prevents the low-level policy from chasing a moving target caused by rapidly changing instruction representations.
\paragraph{\bf Low-Level Training Objective}
We train the low-level executor using an augmented reward signal that combines the environment reward with the predicted reward, 
yielding the following objective:
\begin{equation}
  J_{\pi_{\text{act}}}(\theta_L,\phi_S)
  =
  \mathbb{E}_{\tau\sim \pi_{\text{act}}}
  \left[
	  \sum_{t=0}^{|\tau|-1}\gamma_L^t \left(r_t + \lambda_\text{act} \hat{r}_t\right)
  \right].
  \label{eqa:low-level}
\end{equation}
The augmentation term $\lambda_\text{act} \hat{r}_t$ serves as a dense shaping signal that improves sample efficiency under sparse rewards. Importantly, we still keep the environment reward $r_t$ in the objective to ensure that the learned policy remains grounded in the true task objective, while $\hat{r}_t$ provides additional guidance during training. 
The low-level policy is optimized by maximizing the 
accumulative discounted reward in Eq.~\eqref{eqa:low-level} using the PPO algorithm~\cite{schulman2017proximal}.

\begin{algorithm}[t]
\caption{Hierarchical Reinforcement Learning with Language Instructions}
\label{alg:bilevel}
\begin{algorithmic}[1]
\STATE \textbf{Input:} instruction set $\Ic$, horizon $T$, discount factors $\gamma_L,\gamma_H$
\STATE \textbf{Init:} encoders $\phi_S,\phi_I$, selector $\theta_H$, executor $\theta_L$, buffers $\mathcal{B}_H,\mathcal{B}_L$
\FOR{each epoch}
  \FOR{each episode}
    \STATE Encode all instructions $\{h_i^I\}_{i=1}^{|\Ic|}$ using $f_{\text{inst}}^{\phi_I}$
    \STATE $k\leftarrow 0$
    \WHILE{episode not finished}
      \STATE Compute state embedding $h_{kT}^S=f_{\text{state}}^{\phi_S}(s_{kT})$
      \STATE Compute scores $\ell_{k,i}=\mathrm{score}_{\theta_H}(h_{kT}^S,h_i^I)$ and sample $i_k^*\sim\mathrm{Softmax}(\ell_{k,i})$
      \STATE Set instruction guidance $z_k \leftarrow h_{i_k^*}^I$
      \FOR{$t=kT$ to $(k+1)T-1$}
        \STATE Sample $a_t\sim \pi_{\text{act}}(\cdot|f_{\text{state}}^{\phi_S}(s_t),z_k)$, step env, store $(s_t,a_t,r_t,s_{t+1})$ in $\mathcal{B}_L$
        \STATE \textbf{if} episode ends \textbf{then break}
      \ENDFOR
      \STATE Update reward model $\hat{r}$ using $\mathcal{B}_L$ to minimize Eq.~\eqref{eqa:reward_loss} 
      \STATE Compute predicted segment return $\hat{R}_k=\sum_{t=0}^{T-1}\hat{r}_{kT+t}$
      \STATE Compute auxiliary reward $R_k^\text{aux}$ using Eq.~\eqref{eqa:auxiliary}
      \STATE Store $(s_{kT}, i_k^*, \hat{R}_k, R_k^\text{aux}, s_{(k+1)T})$ in $\mathcal{B}_H$; $k\leftarrow k+1$
    \ENDWHILE
  \ENDFOR
  \STATE Update selector $(\phi_I,\theta_H)$ using $\mathcal{B}_H$ via REINFORCE to maximize Eq.~\eqref{eqa:high-level}
  \STATE Update executor $\theta_L$ using $\mathcal{B}_L$ via PPO to maximize Eq.~\eqref{eqa:low-level}
  \STATE Update state encoder $\phi_S$ using gradients from both 
  (Eq.~\eqref{eqa:high-level} and Eq.~\eqref{eqa:low-level})
\ENDFOR
\end{algorithmic}
\end{algorithm}

%
\subsection{Reward Model}
Language-instructed RL environments often suffer from sparse and delayed rewards, where feedback is only provided after completing a subtask or satisfying a language-conditioned objective. Such sparse supervision makes optimization difficult for both the high-level selector and the low-level executor, particularly during early training when successful trajectories are rarely observed.

To mitigate this issue, we introduce an auxiliary reward model that learns a dense approximation of the environment reward from collected transitions. Concretely, the reward model is implemented as a lightweight MLP head on top of the latent representations produced by the residual CNN state encoder. 
As mentioned in the previous subsections, 
the reward model is used in two ways: (i) to provide step-wise predicted rewards $\hat{r}_t$ for augmenting the executor's training signal; and (ii) to estimate segment-level returns $\hat{R}_k$ for training the selector.
Empirically, we observe that conditioning the reward model on the transition tuple $(s_t,a_t,s_{t+1})$ better captures subtask completion signals than using only $(s_t,a_t)$. This matches the common structure of rewards in language-instructed RL environments, where rewards often depend on state changes rather than static states alone, such as reaching a target location, collecting an item, or triggering a valid interaction specified by the instruction.
Formally, the reward model is defined as
\begin{equation}
\hat{r}_t = \hat{r}_{\psi}(s_t,a_t,s_{t+1}),
\end{equation}
where $\psi$ denotes the reward model parameters and $\hat{r}_t$ is the predicted immediate reward.
The model is trained on transitions sampled from the replay buffer by minimizing the mean squared error (MSE):
\begin{equation}
\mathcal{L}_{\text{RM}}
=
\mathbb{E}_{(s_t,a_t,s_{t+1},r_t)\sim\mathcal{B}_L}
\left[
\left(\hat{r}_{\psi}(s_t,a_t,s_{t+1})-r_t\right)^2
\right],
\label{eqa:reward_loss}
\end{equation}
where $\mathcal{B}_L$ denotes the replay buffer containing transitions collected through low-level policy interactions with the environment.
%
\subsection{HRLLI Algorithm}
Algorithm~\ref{alg:bilevel} summarizes the training procedure of the proposed HRLLI. At each segment $k$, the selector evaluates the state embedding $h_{kT}^S$ and computes relevance scores $\ell_{k,i}$ over instruction embeddings $\{h_i^I\}$, selecting the $i_k^*$-th instruction whose embedding forms the guidance vector $z_k$. Conditioned on $z_k$, the executor interacts with the environment for $T$ steps and stores transitions $(s_t,a_t,r_t,s_{t+1})$ in the low-level buffer $\mathcal{B}_L$. The reward model produces the predicted segment return $\hat{R}_k$, while the auxiliary reward $R_k^{\text{aux}}$
is computed according to Eq.~\eqref{eqa:auxiliary}. The resulting high-level transition $(s_{kT}, i_k^*, \hat{R}_k, R_k^\text{aux}, s_{(k+1)T})$ is then stored in the high-level buffer $\mathcal{B}_H$. 
The reward model is updated by minimizing Eq.~\eqref{eqa:reward_loss}.
The selector, executor, and state encoder are optimized according to 
the objectives in Eq.~\eqref{eqa:high-level} and Eq.~\eqref{eqa:low-level}.
%
\section{Experiments}
\subsection{Experimental Settings}
\subsubsection{Evaluation Benchmark}
We evaluate our approach on the instruction-heavy decision-making benchmark RTFM~\cite{zhong2019rtfm}. RTFM is a text-based gridworld containing a player, monsters, and collectible items placed on a discrete 2D map. The agent receives three types of inputs: a textual state observation describing the gridworld, a short task description specifying the goal, and a wiki document containing instructions on how to complete the task, which we split into sentence-level instructions. The environment is procedurally generated across episodes, with varying instructions and randomized configurations of players, items, and monsters, requiring agents to extract task-relevant knowledge from the document rather than memorize dynamics. The difficulty varies along three dimensions: the number of monsters (One or Two), the language style of task descriptions (SL--Simple Language or NL--Natural Language), 
and the ordering of instruction sentences (fixed or shuffled). We conduct experiments across multiple different settings. 
\subsubsection{Baselines}
We compare our proposed HRLLI against multiple state-of--the-art instruction-based RL methods:
\begin{itemize}
\item \textbf{txt2$\pi$} \cite{zhong2019rtfm}: an instruction-based RL method that models the three-way interactions among the goal, document, and observations.
\item \textbf{Reader} \cite{dainese2023reader}: a model-based language-instructed RL approach that learns a world model of the target environment and performs decision making via MCTS planning over the learned model. It also introduces a Transformer-based baseline using a standard Transformer-based reward model.
\item \textbf{Oracle MCTS}: an upper-bound baseline that assumes access to the ground-truth environment simulator and makes decisions using MCTS planning.
\end{itemize}
\subsubsection{Implementation Details}
Experiments are conducted on the RTFM benchmark~\cite{zhong2019rtfm} using the official Gym environment. The environment is created with room size $10\times10$, and no partial observability by default. The maximum number of candidate instructions is capped at 64. Training runs for $10^6$ environment steps, and evaluation is performed every 5,000 steps using 20 episodes.
\paragraph{Instruction and state encoding.}
Instruction texts are encoded using the pretrained MiniLM encoder (\texttt{sentence-transformers/all-MiniLM-L6-v2})~\cite{wang2020minilm}, producing 384-dimensional normalized sentence embeddings. These embeddings are cached and then projected to 256 dimensions by a two-layer MLP adapter with hidden size 256 and ReLU activation. State encoder is adapted from $\text{txt}2\pi$, where text inputs are encoded with a token embedding layer (dim 64) followed by a bidirectional GRU (hidden size 64 per direction) and attention pooling. The resulting features are combined with observation features and processed by a convolutional encoder with two $3\times3$ layers (64 channels), followed by global max pooling and a linear projection to a 256-dimensional state representation. The task description is encoded together with the state using the state encoder.
\paragraph{Policy architecture and training.}
The high-level instruction selector is implemented as a scoring network with
alternating Linear and ReLU layers, where $d=256$. It scores each candidate instruction conditioned on the current state representation and selects one instruction every $T=10$ steps. 
The high-level selector is updated every 2,000 environment steps with discount factor $\gamma_H=0.99$. 
The low-level controller is trained with PPO using Stable-Baselines3~\cite{raffin2021stable}. Its policy and value networks both use two hidden layers of size 256. The learning rate is $3\times10^{-4}$ and the discount factor is $\gamma_L=0.99$.
The weighting coefficients for reward combinations are set as $\lambda_\text{sel}=0.01$ and $\lambda_\text{act}=0.1$. 
The overall objective used for state encoder update is 
$J_\text{enc}(\phi_S)=\xi_H J_{\pi_{\text{sel}}}(\theta_H,\phi_S,\phi_I)+\xi_L J_{\pi_{\text{act}}}(\theta_L,\phi_S)$,
while the trade-off coefficients $\xi_H=1$ and $\xi_L=0.01$ are used to balance the contributions of the high-level and low-level objectives.

\paragraph{Reward model.}
The reward model is implemented as an MLP with hidden size 256 and alternating Linear and ReLU layers. The replay buffer capacity is 200,000 and the batch size for reward-model training is 256. Each update performs 100 gradient steps using Adam with learning rate $3\times10^{-4}$. 

\subsection{Experimental Results}
\begin{table}[t]
\centering
\caption{Experimental results on the single-monster RTFM task under different language and instruction settings. Results are reported as average win rate (\%) over five runs.
}
\setlength{\tabcolsep}{6pt}
\renewcommand{\arraystretch}{1.15}
\resizebox{\linewidth}{!}{
\begin{tabular}{lcccc}
\toprule
\textbf{Methods}
& \textbf{One SL}
& \textbf{One SL (Shuffle)}
& \textbf{One NL}
& \textbf{One NL (Shuffle)} \\
\midrule
txt2$\pi$ & $0.38 \pm 0.04$ & $0.41 \pm 0.02$ & $0.39 \pm 0.03$ & $0.42 \pm 0.01$ \\
txt2$\pi$ (100M) & \textbf{$0.99 \pm 0.01$} & $0.68 \pm 0.22$ & $0.50 \pm 0.01$ & $0.50 \pm 0.01$ \\
\midrule
Reader & $\mathbf{0.99 \pm 0.01}$ & $0.85 \pm 0.17$ & $0.79 \pm 0.19$ & $0.63 \pm 0.18$ \\
Reader (Transformer) & $0.97 \pm 0.01$ & $0.90 \pm 0.15$ & $0.75 \pm 0.19$ & $0.59 \pm 0.17$ \\
\midrule
	\textbf{HRLLI (Ours)} & $\mathbf{0.99 \pm 0.01}$ & $\mathbf{0.92 \pm 0.19}$ & $\mathbf{0.83 \pm 0.21}$ & $\mathbf{0.79 \pm 0.15}$ \\
\midrule
Oracle MCTS & $0.97 \pm 0.01$ & $0.97 \pm 0.01$ & $0.97 \pm 0.01$ & $0.97 \pm 0.01$ \\
\bottomrule
\end{tabular}
}
\label{tab:single_result}
\end{table}
\begin{table}[t]
\centering
\caption{Experimental results on the two-monster RTFM task under different language and instruction settings. 
Results are reported as average win rate (\%) over five runs.}
\setlength{\tabcolsep}{6pt}
\renewcommand{\arraystretch}{1.15}
\resizebox{\linewidth}{!}{
\begin{tabular}{lcccc}
\toprule
\textbf{Methods}
& \textbf{Two SL}
& \textbf{Two SL (Shuffle)}
& \textbf{Two NL}
& \textbf{Two NL (Shuffle)} \\
\midrule
txt2$\pi$ & $0.09 \pm 0.02$ & $0.09 \pm 0.02$ & $0.07 \pm 0.01$ & $0.08 \pm 0.02$ \\
txt2$\pi$ (100M) & $0.24 \pm 0.00$ & $0.24 \pm 0.00$ & $0.24 \pm 0.01$ & $0.24 \pm 0.01$ \\
\midrule
Reader & $0.50 \pm 0.07$ & $0.47 \pm 0.16$ & $0.24 \pm 0.01$ & $0.24 \pm 0.01$ \\
Reader (Transformer) & $0.50 \pm 0.07$ & $0.39 \pm 0.10$ & $0.23 \pm 0.03$ & $0.23 \pm 0.01$ \\
\midrule
	\textbf{HRLLI (Ours)}
& $\mathbf{0.60 \pm 0.08}$ & $\mathbf{0.55 \pm 0.06}$ & $\mathbf{0.30 \pm 0.03}$ & $\mathbf{0.29 \pm 0.02}$ \\
\midrule
Oracle MCTS
& $0.83 \pm 0.01$
& $0.83 \pm 0.01$
& $0.83 \pm 0.01$
& $0.83 \pm 0.01$ \\
\bottomrule
\end{tabular}
}
\label{tab:two_result}
\vskip -0.1in
\end{table}
Tables~\ref{tab:single_result} and~\ref{tab:two_result} report the experimental results on the single-monster and two-monster RTFM tasks under different language and instruction settings. 
The proposed method consistently achieves the best performance across different settings, demonstrating the effectiveness of explicitly modeling instruction selection in language-instructed RL.
On the simpler single-monster setting (Table~\ref{tab:single_result}), most approaches are able to achieve relatively strong performance when structured language instructions are available. However, performance degrades noticeably under shuffled or natural language settings, indicating that existing methods remain sensitive to irrelevant or poorly aligned instructions. In contrast, our method maintains strong and stable performance across all settings, achieving the highest win rates in every configuration. This suggests that dynamically selecting state-relevant instructions allows the agent to better utilize useful guidance while avoiding distraction from irrelevant instructions.
The advantage of our approach becomes more evident in the more challenging two-monster environment shown in Table~\ref{tab:two_result}. Compared with Reader and its Transformer baseline, which condition the policy on the entire instruction set, our hierarchical framework achieves consistent improvements across all language settings. In particular, the performance gain is more pronounced in natural language scenarios, where instructions are less structured and contain higher variability. These results indicate that treating instructions as state-dependent instructions and explicitly learning when to use them significantly improves decision making under complex environments.
Furthermore, while Oracle MCTS achieves the highest performance by assuming the MCTS has access to the ground-truth simulator, our method substantially narrows the gap without relying on environment planning or privileged information. This highlights the benefit of separating instruction reasoning and action execution through the proposed hierarchical optimization framework.
\subsection{Ablation Study}
\begin{table}[t]
\centering
\caption{Ablation study on different variants in the Two NL setting. Results are reported as win rate (\%) and averaged over three runs.}
\setlength{\tabcolsep}{6pt}
\renewcommand{\arraystretch}{1.15}
\resizebox{0.72\linewidth}{!}{
\begin{tabular}{lcc}
\toprule
\textbf{Methods}
& \textbf{Two NL}
& \textbf{Two NL (Shuffle)} \\
\midrule
HRLLI (Full method) & $0.30\pm0.03$ & $0.29\pm0.02$ \\
w/o hard selection & $0.29\pm0.03$ & $0.28\pm0.04$ \\
w/o high-level encoder update & $0.24\pm0.05$ & $0.25\pm0.03$ \\
value-diff auxiliary & $0.27\pm0.02$ & $0.27\pm0.03$ \\
\bottomrule
\end{tabular}
}
\label{tab:ablation}
\end{table}
We conduct an ablation study to evaluate several variants of our HRLLI method. Specifically, we compare the full HRLLI method with three variants: 
\textit{w/o hard selection}, where instructions are sampled from the high-level policy during evaluation rather than selected via hard selection;
(2) \textit{w/o high-level encoder update}, which updates the shared state encoder only through the low-level objective, 
without gradients from the high-level policy objective (i.e., $\xi_H = 0$); 
(3) \textit{value-diff auxiliary}, which replaces the cosine-similarity auxiliary reward with a value-difference reward 
$R_k^{\text{aux}} = V_L(s_{(k+1)T}) - V_L(s_{kT})$.
The results are summarized in Table~\ref{tab:ablation}. 
We observe that replacing hard selection with sampling in the evaluation slightly degrades the performance in the \textit{w/o hard selection} variant. 
When the state encoder is updated only through the low-level policy (\textit{w/o high-level encoder update}), the performance drops significantly, indicating that the shared state encoder benefits from joint updates from both high-level and low-level policies. 
The \textit{value-diff auxiliary} variant also consistently reduces performance on both Two NL and Two NL (Shuffle), suggesting that the cosine-similarity auxiliary reward provides more effective guidance for training the high-level instruction policy, although the value-difference reward can still provide a reasonable learning signal.
\begin{figure}[t]
\centering
\includegraphics[width=0.95\linewidth]{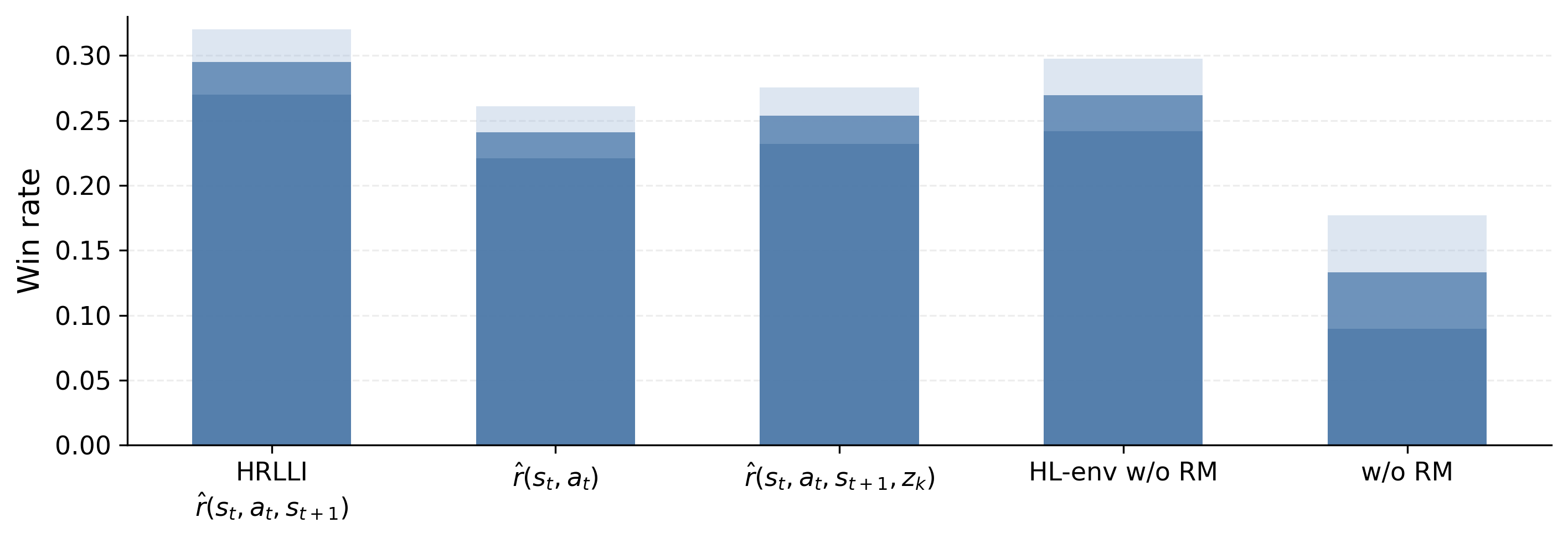}
\vskip -0.1in
\caption{
Ablation study on different reward designs in the Two NL (Shuffle) setting.  Bars show the mean performance (Win rate) and the shaded area indicates the standard deviation. Results are averaged over three runs.
}
\label{fig:ablation}
\vskip -0.1in
\end{figure}
We further conduct an ablation study on the design and utilization of the reward model in the Two NL (Shuffle) setting.
We compare the following variants: 
(1) $\text{HRLLI}~\hat{r}(s_t,a_t,s_{t+1})$, our full method that learns a transition-based reward model; 
(2) $\hat{r}(s_t,a_t)$, which learns a reward model conditioned only on the state–action pair $(s_t,a_t)$; 
(3) $\hat{r}(s_t,a_t,s_{t+1},z_k)$, which additionally conditions the reward model on the instruction guidance $z_k$; 
(4) \textit{HL-env w/o RM}, which removes the reward model from high-level policy learning and instead uses the environment reward; 
(5) \textit{w/o RM}, which removes the reward model entirely from both high-level and low-level policy learning.
The results are summarized in Figure~\ref{fig:ablation}. 
From the figure, we observe that the full HRLLI method achieves the best performance, demonstrating the effectiveness of the proposed reward model design. 
The reward model performs best when conditioned on the transition $(s_t,a_t,s_{t+1})$, which is consistent with the sparse-reward setting of the environment. 
Conditioning the reward model on the instruction guidance $z_k$ does not provide additional performance gains. 
Interestingly, replacing the reward model with the environment reward in the high-level policy learning (\textit{HL-env w/o RM}) does not significantly degrade the performance. 
We hypothesize that the high-level policy already captures long-term signals through the cumulative segment reward, while the reward model provides additional guidance that leads to further performance gains.
In contrast, removing the reward model entirely (\textit{w/o RM}) leads to a significant drop in win rate, indicating that the reward model plays a crucial role in low-level policy learning. 
Without a well-trained reward model, the low-level policy cannot effectively learn from the sparse reward signal. 
The performance degradation in the variants $\hat{r}(s_t,a_t)$ and $\hat{r}(s_t,a_t,s_{t+1},z_k)$ further highlights the importance of an appropriate reward model design for low-level policy learning.
\subsection{Hyperparameter Sensitivity Analysis}
\begin{figure}[t]
\centering
\includegraphics[width=0.48\linewidth]{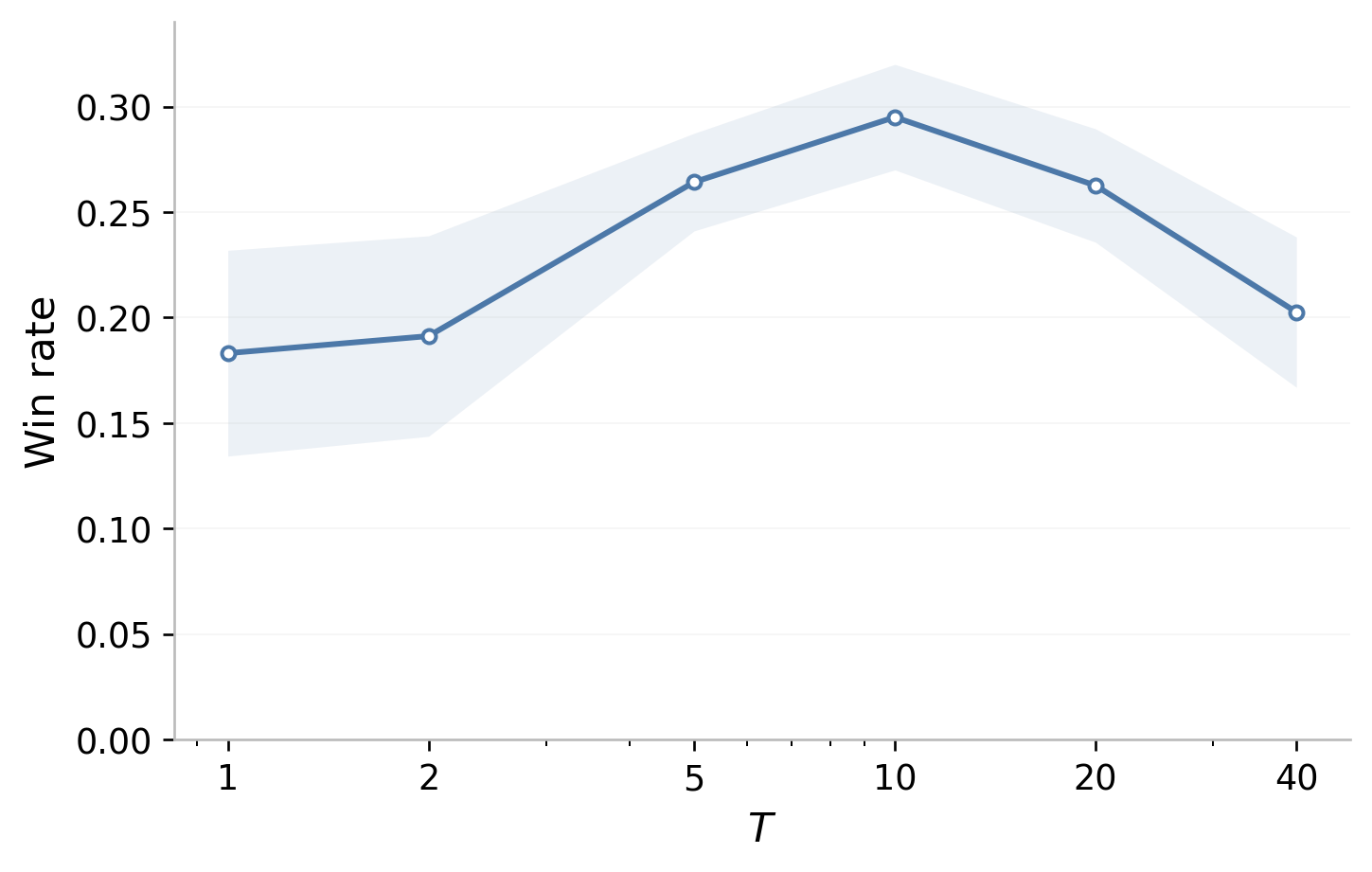}
\hfill
\includegraphics[width=0.48\linewidth]{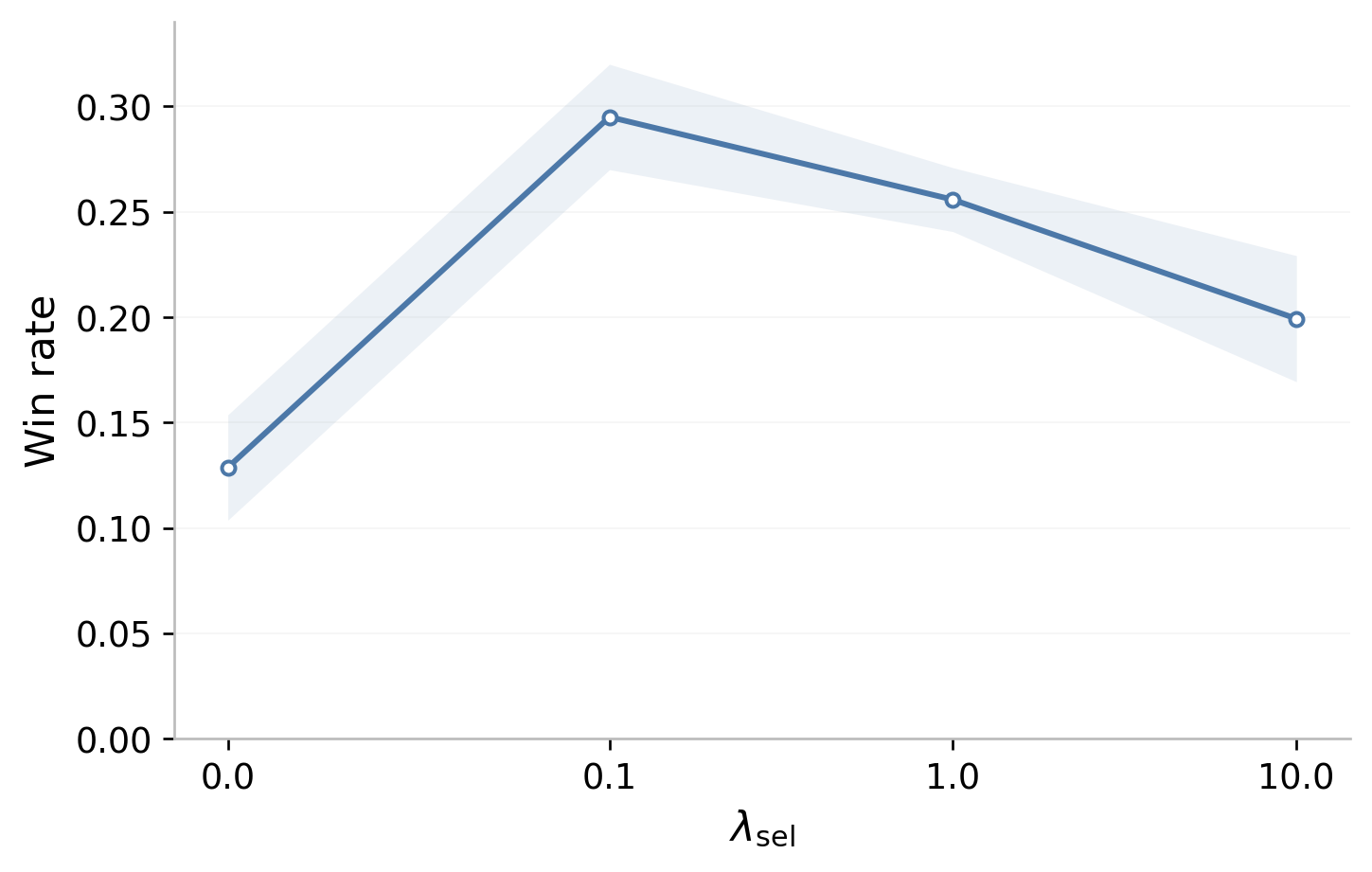}
\caption{
Sensitivity analysis of key hyperparameters in the Two NL (Shuffle) setting.
\textbf{Left:} sensitivity to the high-level horizon $T$.
\textbf{Right:} sensitivity to the auxiliary reward weight $\lambda_\text{sel}$.
Results are averaged over three runs.
}
\label{fig:sensitivity}
\vskip -0.1in
\end{figure}
We further conduct a hyperparameter sensitivity analysis on the high-level horizon $T$ and the auxiliary reward weight $\lambda_{\text{sel}}$ on the Two NL (Shuffle) setting to evaluate how our method behaves under different choices of key parameters.
The results are summarized in Figure~\ref{fig:sensitivity}. From the left panel, we observe that the choice of horizon $T$ has a moderate impact on the overall win rate. In particular, values of $T$ between 5 and 20 lead to similar performance. When the horizon becomes too large ($T=40$), the selected instruction remains unchanged for a long duration, which reduces the flexibility of guidance and slightly degrades performance. On the other hand, when $T=1$, the method effectively degrades to a standard instruction-conditioned policy where a new instruction is selected at every timestep. In this case, the hierarchical structure collapses into a single-level policy, leading to noticeably worse performance and higher variance. This suggests that overly frequent changes in instruction guidance may destabilize the decision process. This result demonstrates the effectiveness of our hierarchical design, where a high-level instruction policy provides short-horizon semantic guidance that improves the decision making of the low-level action policy.
From the right panel, we observe that the trade-off parameter $\lambda_{\text{sel}}$ also affects performance. As $\lambda_{\text{sel}}$ increases, the overall win rate decreases, suggesting that an overly strong auxiliary reward may dominate the learning signal and negatively impact training. However, removing the auxiliary reward entirely ($\lambda_{\text{sel}}=0$) also leads to a clear performance drop. This indicates that the auxiliary signal provides useful guidance during training, while a moderate weight achieves the best overall performance, indicating that our chosen value provides an effective balance between the environment reward and the auxiliary objective.
%

%
\section{Conclusion}
In this paper, we study a relatively underexplored setting of language-instructed RL, where an RL agent can exploit external language knowledge provided as a set of instructions. Unlike standard instruction-conditioned policies that consume the entire instruction globally, our setting requires the agent to determine which parts of the available language instructions are relevant at different stages of interaction. To address this challenge, we propose \textbf{H}ierarchical \textbf{R}einforcement \textbf{L}earning with \textbf{L}anguage \textbf{I}nstructions (HRLLI), a hierarchical selector–executor framework that treats language instructions as high-level semantic guidance and uses them to guide low-level policy learning. Experiments show that HRLLI consistently outperforms state-of-the-art language-instructed RL methods on the strong instruction-conditioned RTFM benchmarks, particularly in more challenging natural-language and shuffled-document settings. These results suggest that, in instruction-heavy RL environments, effective use of language depends not only on multimodal grounding, but also on explicit control over \textit{which} instruction to use and \textit{when} to use it.
%
%
%
\bibliographystyle{splncs04}
\bibliography{refs}

@inproceedings{zhangECAI25,
  title={Skill-Enhanced Reinforcement Learning Acceleration from Heterogeneous Demonstrations},
  author={Hanping Zhang and Yuhong Guo},
  booktitle={European Conference on Artificial Intelligence (ECAI)},
  year={2025}
}

@inproceedings{zhong2019rtfm,
  title={{RTFM}: Generalising to Novel Environment Dynamics via Reading},
  author={Zhong, Victor and Rockt{\"a}schel, Tim and Grefenstette, Edward},
  booktitle={International Conference on Learning Representations (ICLR)},
  year={2020}
}

@inproceedings{dainese2023reader,
  title={Reader: Model-based language-instructed reinforcement learning},
  author={Dainese, Nicola and Marttinen, Pekka and Ilin, Alexander},
  booktitle={Conference on Empirical Methods in Natural Language Processing (EMNLP)},
  year={2023}
}

@inproceedings{shridhar2020alfred,
  title={{ALFRED}: A Benchmark for Interpreting Grounded Instructions for Everyday Tasks},
  author={Shridhar, Mohit and Thomason, Jesse and Gordon, Daniel and Bisk, Yonatan and Han, Winson and Mottaghi, Roozbeh and Zettlemoyer, Luke and Fox, Dieter},
  booktitle={IEEE/CVF Conference on Computer Vision and Pattern Recognition (CVPR)},
  year={2020}
}

@inproceedings{shridhar2020alfworld,
  title ={{ALFWorld}: Aligning Text and Embodied Environments for Interactive Learning},
  author={Mohit Shridhar and Xingdi Yuan and Marc-Alexandre C\^ot\'e and Yonatan Bisk and Adam Trischler and Matthew Hausknecht},
  booktitle = {International Conference on Learning Representations (ICLR)},
  year = {2021}
}

@inproceedings{ouyang2022training,
  title={Training language models to follow instructions with human feedback},
  author={Ouyang, Long and Wu, Jeffrey and Jiang, Xu and Almeida, Diogo and Wainwright, Carroll and Mishkin, Pamela and Zhang, Chong and Agarwal, Sandhini and Slama, Katarina and Ray, Alex and others},
  booktitle={Advances in Neural Information Processing Systems (NeurIPS)},
  year={2022}
}

@book{sutton1998reinforcement,
  title={Reinforcement Learning: An Introduction},
  author={Sutton, Richard S. and Barto, Andrew G.},
  publisher={MIT Press},
  year={1998}
}

@inproceedings{luketina2019survey,
  title={A Survey of Reinforcement Learning Informed by Natural Language}, 
  author={Jelena Luketina and Nantas Nardelli and Gregory Farquhar and Jakob Foerster and Jacob Andreas and Edward Grefenstette and Shimon Whiteson and Tim Rocktäschel},
  booktitle={International Joint Conference on Artificial Intelligence (IJCAI)},
  year={2019}
}

@article{tang2025deep,
  title={Deep Reinforcement Learning for Robotics: A Survey of Real-World Successes},
  author={Tang, Chen and Abbatematteo, Ben and Hu, Jiaheng and Chandra, Rohan and Mart{\'\i}n-Mart{\'\i}n, Roberto and Stone, Peter},
  journal={Annual Review of Control, Robotics, and Autonomous Systems (ARCRAS)},
  year={2025}
}

@misc{berner2019dota,
  title={Dota 2 with Large Scale Deep Reinforcement Learning},
  author={OpenAI and Christopher Berner and Greg Brockman and Brooke Chan and Vicki Cheung and Przemysław Dębiak and Christy Dennison and David Farhi and Quirin Fischer and Shariq Hashme and Chris Hesse and Rafal Józefowicz and Scott Gray and Catherine Olsson and Jakub Pachocki and Michael Petrov and Henrique Pondé de Oliveira Pinto and Jonathan Raiman and Tim Salimans and Jeremy Schlatter and Jonas Schneider and Szymon Sidor and Ilya Sutskever and Jie Tang and Filip Wolski and Susan Zhang},
  archivePrefix={arXiv},
  eprint={1912.06680},
  year={2019}
}

@article{kiran2021deep,
  title={Deep Reinforcement Learning for Autonomous Driving: A Survey},
  author={Kiran, B Ravi and Sobh, Ibrahim and Talpaert, Victor and Mannion, Patrick and Al Sallab, Ahmad A and Yogamani, Senthil and P{\'e}rez, Patrick},
  journal={IEEE Transactions on Intelligent Transportation Systems (T-ITS)},
  year={2021}
}

@inproceedings{hu2023language,
  title={Language Instructed Reinforcement Learning for Human-AI Coordination},
  author={Hu, Hengyuan and Sadigh, Dorsa},
  booktitle={International Conference on Machine Learning (ICML)},
  year={2023}
}

@inproceedings{pertsch2021accelerating,
  title={Accelerating Reinforcement Learning with Learned Skill Priors},
  author={Pertsch, Karl and Lee, Youngwoon and Lim, Joseph},
  booktitle={Conference on Robot Learning (CoRL)},
  year={2021}
}

@inproceedings{wang2020minilm,
  title={{MiniLM}: Deep Self-Attention Distillation for Task-Agnostic Compression of Pre-Trained Transformers},
  author={Wang, Wenhui and Wei, Furu and Dong, Li and Bao, Hangbo and Yang, Nan and Zhou, Ming},
  booktitle={Advances in Neural Information Processing Systems (NeurIPS)},
  year={2020}
}

@article{mees2022calvin,
  title={{CALVIN}: A Benchmark for Language-Conditioned Policy Learning for Long-Horizon Robot Manipulation Tasks},
  author={Mees, Oier and Hermann, Lukas and Rosete-Beas, Erick and Burgard, Wolfram},
  journal={IEEE Robotics and Automation Letters (RA-L)},
  year={2022}
}

@article{raffin2021stable,
  title={{Stable-Baselines3}: Reliable Reinforcement Learning Implementations},
  author={Raffin, Antonin and Hill, Ashley and Gleave, Adam and Kanervisto, Anssi and Ernestus, Maximilian and Dormann, Noah},
  journal={Journal of Machine Learning Research (JMLR)},
  year={2021}
}

@inproceedings{yu2020meta,
  title={{Meta-World}: A Benchmark and Evaluation for Multi-Task and Meta Reinforcement Learning},
  author={Yu, Tianhe and Quillen, Deirdre and He, Zhanpeng and Julian, Ryan and Hausman, Karol and Finn, Chelsea and Levine, Sergey},
  booktitle={Conference on Robot Learning (CoRL)},
  year={2020}
}

@inproceedings{sutton1999policy,
  title={Policy Gradient Methods for Reinforcement Learning with Function Approximation },
  author={Sutton, Richard S and McAllester, David and Singh, Satinder and Mansour, Yishay},
  booktitle={Advances in Neural Information Processing Systems (NeurIPS)},
  year={1999}
}

@misc{schulman2017proximal,
  title={Proximal Policy Optimization Algorithms},
  author={Schulman, John and Wolski, Filip and Dhariwal, Prafulla and Radford, Alec and Klimov, Oleg},
  archivePrefix={arXiv},
  eprint={1707.06347},
  year={2017}
}

@inproceedings{yan2024efficient,
  title={Efficient Reinforcement Learning with Large Language Model Priors},
  author={Yan, Xue and Song, Yan and Feng, Xidong and Yang, Mengyue and Zhang, Haifeng and Ammar, Haitham Bou and Wang, Jun},
  booktitle={International Conference on Learning Representations (ICLR)},
  year={2025}
}

@inproceedings{kocsis2006bandit,
  title={Bandit based Monte-Carlo Planning},
  author={Kocsis, Levente and Szepesv{\'a}ri, Csaba},
  booktitle={European Conference on Machine Learning (ECML)},
  year={2006}
}

@inproceedings{baek2025ipcgrl,
  title={{IPCGRL}: Language-Instructed Reinforcement Learning for Procedural Level Generation},
  author={Baek, In-Chang and Kim, Sung-Hyun and Lee, Seo-Young and Kim, Dong-Hyeon and Kim, Kyung-Joong},
  booktitle={IEEE Conference on Games (CoG)},
  year={2025}
}

@inproceedings{wang2024llm,
  title = {{LLM}-Empowered State Representation for Reinforcement Learning},
  author = {Wang, Boyuan and Qu, Yun and Jiang, Yuhang and Shao, Jianzhun and Liu, Chang and Yang, Wenming and Ji, Xiangyang},
  booktitle = {International Conference on Machine Learning (ICML)},
  year = {2024}
}
\end{document}